\documentclass[a4paper,fleqn]{cas-dc}

\usepackage[numbers]{natbib}
\usepackage{caption}
\usepackage{pifont}
\usepackage{xcolor} % 确保导言区已加载
\definecolor{rank1}{RGB}{255,0,0}   % 定义排名颜色（红色）
\definecolor{rank2}{RGB}{0,0,255}   % 蓝色
\definecolor{rank3}{RGB}{0,128,0}   % 绿色
\def\tsc#1{\csdef{#1}{\textsc{\lowercase{#1}}\xspace}}
\tsc{WGM}
\tsc{QE}
\tsc{EP}
\tsc{PMS}
\tsc{BEC}
\tsc{DE}

\begin{document}
\let\WriteBookmarks\relax
\def\floatpagepagefraction{1}
\def\textpagefraction{.001}

\shorttitle{CEBSNet}

\shortauthors{Q. Xu et~al.}

% \title [mode = title]{CEBSNet: Change Detection with Change-related Feature Excitation and Background Feature Suppression}                
\title [mode = title]{CEBSNet: Change-Excited and Background-Suppressed Network with Temporal Dependency Modeling for Bitemporal Change Detection}

% \tnotetext[1]{This research was funded by the Scientific Research Program of Tianjin Municipal Education Commission 2023KJ232 and the Tianjin Natural Science Foundation General Project 24JCYBJC00990.}

% \tnotetext[2]{The second title footnote which is a longer text matter
   % to fill through the whole text width and overflow into
   % another line in the footnotes area of the first page.}

\author[1]{Qi’ao Xu}[type=editor, auid=000, orcid=0009-0002-5567-438X]
\fnmark[1]

\ead{qaxu00@163.com}

%\ead[url]{www.cvr.cc, cvr@sayahna.org}

\credit{Conceptualization of this study, Methodology, Software}

\affiliation[1]{organization={College of Computer Science and Technology, Civil Aviation University of China},
    city={Tianjin},
    postcode={300300}, 
    country={China}}

\author[2]{Yan Xing}[style=chinese]
\fnmark[1]
\ead{yxing@cauc.edu.cn}

\author[1]{Jiali Hu}[style=chinese]
% \ead{cvr3@sayahna.org}
% \ead[URL]{www.sayahna.org}

\credit{Data curation, Writing - Original draft preparation}

\affiliation[2]{organization={College of Safety Science and Engineering, Civil Aviation University of China},
    city={Tianjin},
    postcode={300300}, 
    state={},
    country={China}}

\author[1]{Yunan Jia}[style=chinese]
% \fnmark[1,3]
% \ead{rishi@stmdocs.in}
% \ead[URL]{www.stmdocs.in}

\author[1]{Rui Huang}[style=chinese]
\ead{rhuang@cauc.edu.cn}
\cormark[1]

% \cortext[cor1]

\cortext[cor1]{Corresponding author: Rui Huang}
% \cortext[cor2]{Principal corresponding author}

% % Footnote text
\fntext[1]{These authors contributed equally to this work.}  % 共一说明

% \fntext[fn1]{This is the first author footnote. but is common to third
% \fntext[fn2]{Another author footnote, this is a very long footnote and
%   it should be a really long footnote. But this footnote is not yet
%   sufficiently long enough to make two lines of footnote text.}

% % For a title note without a number/mark
% \nonumnote{This note has no numbers. In this work we demonstrate $a_b$
%   the formation Y\_1 of a new type of polariton on the interface
%   between a cuprous oxide slab and a polystyrene micro-sphere placed

\begin{abstract}
Change detection, a critical task in remote sensing and computer vision, aims to identify pixel-level differences between image pairs captured at the same geographic area but different times. It faces numerous challenges such as illumination variation, seasonal changes, background interference, and shooting angles, especially with a large time gap between images. While current methods have advanced, they often overlook temporal dependencies and overemphasize prominent changes while ignoring subtle but equally important changes. To address these limitations, we introduce \textbf{CEBSNet}, a novel change-excited and background-suppressed network with temporal dependency modeling for change detection. During the feature extraction, we utilize a simple Channel Swap Module (CSM) to model temporal dependency, reducing differences and noise. The Feature Excitation and Suppression Module (FESM) is developed to capture both obvious and subtle changes, maintaining the integrity of change regions. Additionally, we design a Pyramid-Aware Spatial-Channel Attention module (PASCA) to enhance the ability to detect change regions at different sizes and focus on critical regions. We conduct extensive experiments on three common street view datasets and two remote sensing datasets, and our method achieves the state-of-the-art performance.
\end{abstract}


\maketitle

\section{Introduction}
Change detection (CD) focuses on detecting pixel-level changes in image pairs captured at the same location but different times \cite{asokan2019change}. These techniques have widespread applications in various fields, such as disaster monitoring and assessment, urban planning and development, natural resource monitoring and utilization, farmland surveys, traffic regulation, and remote sensing~\citep{cheng2024change,Holail2024nsr}. As sensor technology advances, the complexity of image pairs has notably increased. Traditional methods mostly rely on image information like gray-scale/color, texture, and shape, which makes it difficult to handle complex noise interferences and severely limits detection performance~\citep{taneja2011image}. Recently, deep learning-based CD methods have gained widespread attention due to their robust feature representation capabilities, which effectively reducing noise interference~\citep{bai2023deep}.

\begin{figure*}[!htb]
    \centering
    \includegraphics[width=0.8\textwidth]{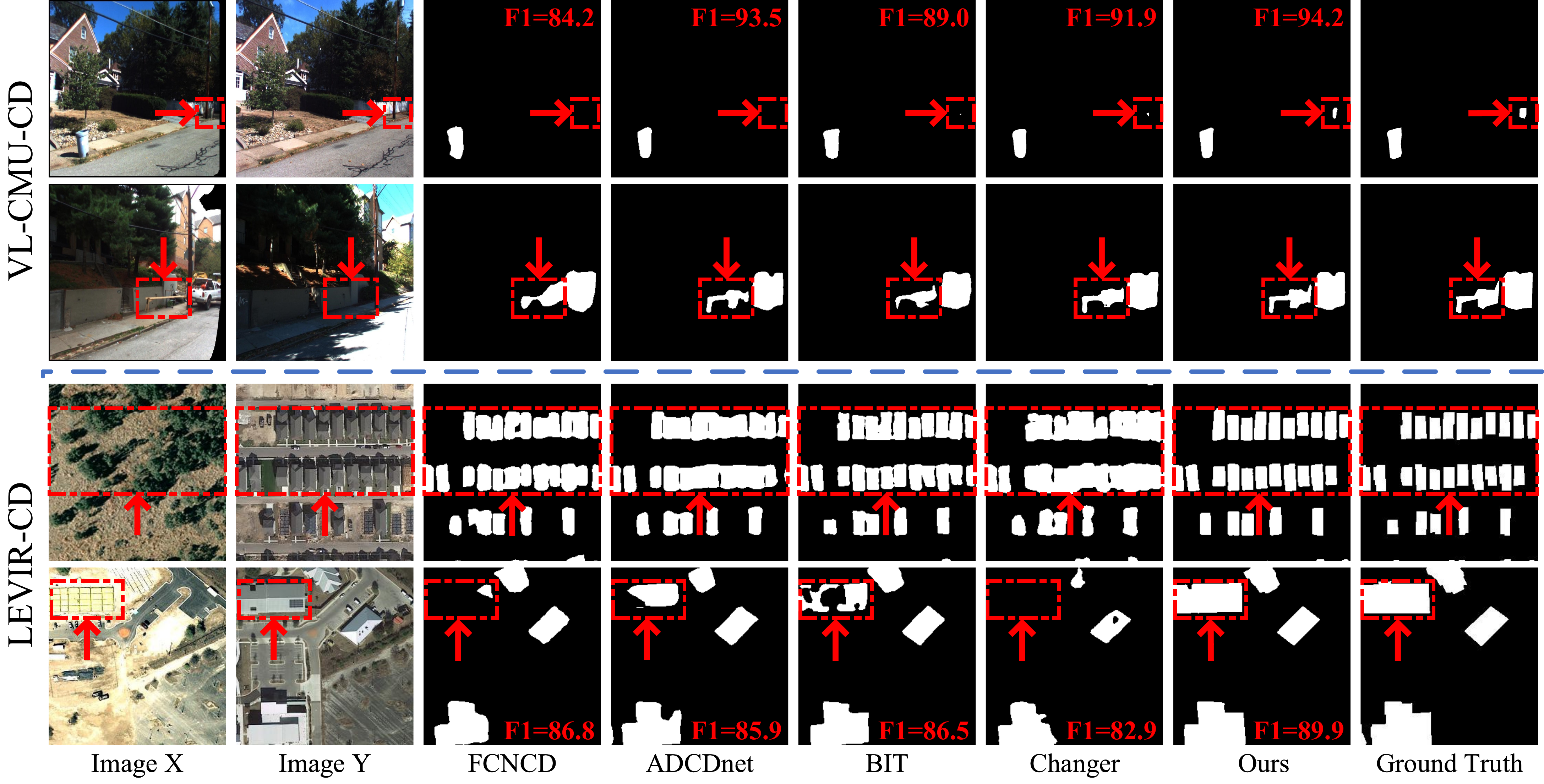}
    \caption{Motivation analysis. We compare our CEBSNet with existing CD methods on VL-CMU-CD and LEVIR-CD datasets. The objects in \textcolor{red}{red dotted box} are important change regions, and the \textcolor{red}{F1 scores} objectively measure the performance.}
    \label{fig:motivation}
\end{figure*}

Despite the notable progress, current CD methods still face critical challenges when applied to complex scenarios. The primary challenge stems from the substantial noise interference caused by temporal differences. As shown in Figure~\ref{fig:motivation}, this limitation manifests in two key aspects: 
1) \textbf{Susceptibility to various noise}: Image pairs are highly vulnerable to factors such as illumination variation, seasonal changes, background interference, and shooting angles, leading to reduced detection precision and limited cross-scenario adaptability.
2) \textbf{Neglect of subtle changes}: Most models focus on detecting obvious changes while overlooking subtle changes, resulting in missed detections and constrained minor-change detection capability.

We contend that the underlying reasons for the subpar performance are the low-quality change features, arising from two main factors: 
1) \textbf{Temporal dependency neglect}: Siamese networks treat image pairs as temporally independent entities, failing to capture latent temporal dependencies. Significant inter-pair discrepancies lead to noise susceptibility and inaccurate change detection. 
2) \textbf{Imbalanced focus on change regions}: Excessive attention to prominent changes causes neglect of subtle yet critical changes, substantially increasing missed detection rates.
Modeling the temporal dependency between image pairs seems promising. It can reduce noise and differences, improving detection performance. But it may weaken the distinct changes between pairs and still miss subtle changes due to the long-standing bias.

To tackle the aforementioned issues, we present \textbf{CEBSNet}, a novel change-excited and background-suppressed network with temporal dependency modeling for change detection. It utilizes a simple Channel Swap Module (CSM) to model the potential temporal dependency between image pairs during feature extraction, alleviating differences and noise interference. The Feature Excitation and Suppression Module (FESM) capture both prominent and subtle changes by exciting change-related regions and suppressing background noise. Moreover, a Pyramid-Aware Spatial-Channel Attention module (PASCA) is developed. By leveraging multi-scale receptive fields and attention mechanisms, this combination effectively boosts the ability to detect change objects at various scales. We conduct extensive experiments on five CD datasets, and our method achieves the state-of-the-art performance.

Our contributions are summarized as follows:
\begin{itemize}
    \item We present an advanced change detection method CEBSNet. It applies a Channel Swap Module (CSM) to explore the temporal correlation between image pairs during feature extraction.
    \item We develop a Feature Excitation and Suppression Module (FESM) to capture both obvious and subtle changes. Additionally, a Pyramid-Aware Spatial-Channel Attention module (PASCA) is designed to aggregates change regions at various sizes.
    \item Extensive experiments conducted on three common street-view CD datasets and two remote sensing CD datasets demonstrate the superiority and effectiveness of our CEBSNet, validating its practical value.
\end{itemize}

The subsequent sections of this paper are organized as follows: Section~\ref{relatedwork} reviews related work on change detection along with feature excitation and suppression. Section~\ref{methodology} introduces the proposed CEBSNet. Section~\ref{experiment} presents the experimental results. Section~\ref{conclusion} concludes with a summary.

\section{Related Work}
\label{relatedwork}

\subsection{Change Detection}

In recent years, change detection has gained increasing attention in computer vision. Traditional methods like Support Vector Machines (SVM), Principal Component Analysis (PCA), Gaussian Mixture Models (GMM), K-means clustering, due to their limitations, struggle to meet the current demands for high-speed and accurate detection~\cite{shi2020change}. 

The development and application of deep-learning technologies have revolutionized change detection. These methods, with strong learning and representation capabilities, have improved performance significantly~\cite{shafique2022deep}. For example, Daudt et al.~\cite{daudt2018fully} introduces three structures: early-fusion EC-EF, mid-fusion FC-Siam-conc and FC-Siam-diff. Huang et al.~\cite{huang2020change} presents ADCDNet, a multi-scale change detection network that uses a siamese network to extract features and compute absolute differences for change detection. Bandara et al.~\cite{bandara2022transformer} proposes ChangeFormer, a siamese network with a hierarchical transformer encoder. It fuses multi-scale difference features to detect changes. Fang et al.~\cite{fang2023changer} proposes Changer, which processes image pairs through an encoder's interaction-layer sequence and uses flow-based dual-alignment fusion for feature alignment. Lv et al.~\cite{lv2023multiscale} introduces a method integrating a multi-scale attention network and change-gradient images to fuse multi-scale features for change detection. Zhao et al.~\cite{zhao2024multiscale} introduces MAPNet, a network that performs multi-scale alignment and progressive feature fusion to enhance detection performance.

Despite these advancements, most methods do not consider temporal correlation between images when extracting features. So they are easily affected by background noise. Although some methods have tried image concatenation or feature-pair fusion, the feature interaction remains insufficient. To tackle this issue, this paper proposes a simple and effective channel swap module. By exchanging feature information between image pairs, it explores their potential temporal correlation, reduces background differences and noise interference, and improves change detection accuracy.

\subsection{Feature Excitation and Suppression}

Feature excitation and suppression play a crucial role in deep learning by enhancing feature responses in regions of interest while suppressing irrelevant or redundant information. The attention mechanism represents the most prevalent means of achieving this objective. It has been widely applied in various fields such as natural language processing, computer vision, and time series analysis~\cite{niu2021review}. At its essence, the attention mechanism entails the learning of a set of weight coefficients. By applying these weights to features, it foregrounds crucial features and diminishes the influence of unimportant ones. For instance, SENet~\cite{hu2018squeeze} adjusts channel importance by modeling channel interdependencies, improving the network's representation ability. Self-Attention~\cite{vaswani2017attention} focuses on modeling global feature correlations, capturing long-range dependencies to make up for CNNs' focus on local information. DANet~\cite{fu2019dual} and CBAM~\cite{woo2018cbam} adjusts spatial and channel weights in both dimensions to highlight important features. 

In change detection, these processes are assume a critical role. For example, STANet~\cite{chen2020spatial} designs a change self-attention mechanism to calculate weighted spatio-temporal dependencies for feature excitation. DMINet~\cite{feng2023change} adopts a dual-branch multi-layer spatio-temporal network with self-attention and cross-attention integration to enhance change features, addressing the issue of foreground-background imbalance. A2Net~\cite{li2023lightweight} uses progressive feature aggregation and a attention module to identify changes and enhances feature representation. Transformer-based methods like BIT~\cite{chen2021remote}, ACAHNet~\cite{zhang2023asymmetric}, and M-Swin~\cite{pan2024m} primarily utilize self-attention for feature extraction or spatial information modeling to highlight important change regions.

\begin{figure*}[tb]
    \centering
    \includegraphics[width=0.85\textwidth]{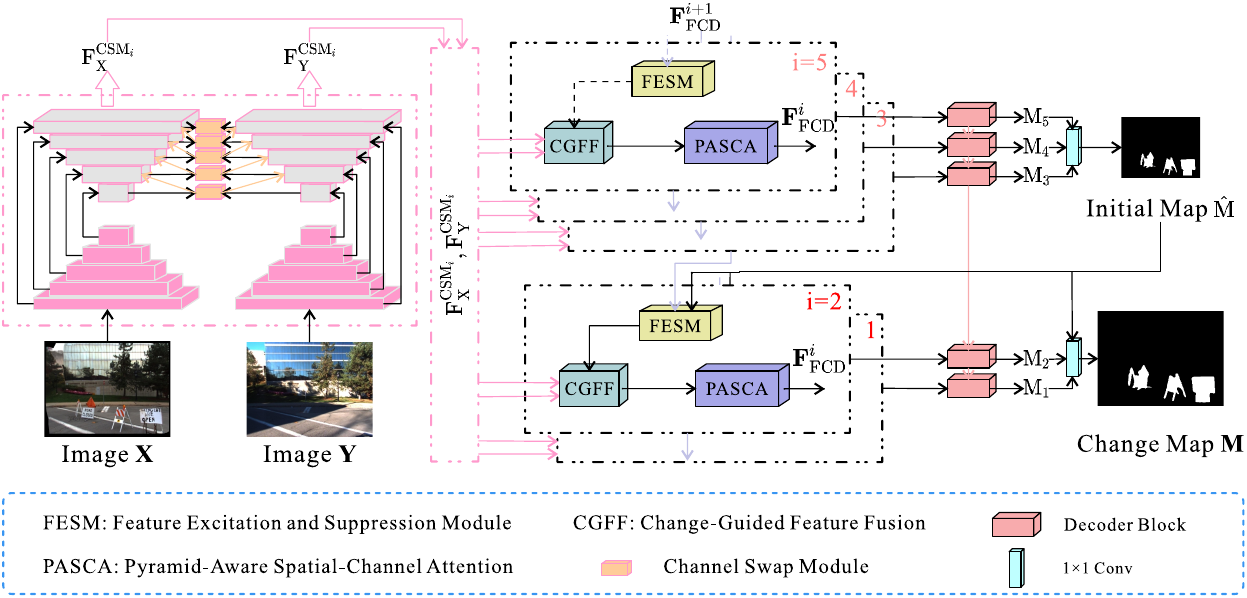}
    \caption{The architecture of our CEBSNet. It comprises three key stages: 1) Spatial-Temporal Feature Extraction, where a siamese encoder with Channel Swap Module (CSM) extracts multi-scale spatiotemporal feature pairs; 2) Change Feature Refinement, where change features are generated and iteratively refined through CGFF, FESM and PASCA; and 3) Multi-Scale Change Detection, where refined change features are decoded into multi-scale change maps and fused to generate the final change map.}
    \label{fig:framework}
\end{figure*}

Current CD methods mainly focus on obvious change regions, often overlook less prominent but important change regions. To address this issue, this paper put forward a dual-branch feature excitation and suppression module. This module helps the network learn both obvious and subtle changes, resulting in more accurate change regions. It also solves the problem of reduced differences when modeling image temporal correlation, improving detection accuracy.

\section{Methodology}
\label{methodology}

\subsection{Overview}
As shown in Figure~\ref{fig:framework}, our CEBSNet consists of three main stages: Spatial-Temporal Feature Extraction, Change Feature Refinement, and Multi-Scale Change Detection. A siamese encoder with shared weights processes the image pair $\textbf{X}$ and $\textbf{Y}$, generating five pairs of multi-scale features $ \{ \text{F}_\text{X}^{\text{CSM}_i}, \text{F}_\text{Y}^{\text{CSM}_i} \}^5_{i=1}$ that encode hierarchical spatial-semantic information. Multi-scale feature pairs undergo iterative refinement through feature excitation and attention modules. The refined change features $\{\text{F}_\text{FCD}^i \}^5_{i=1}$ are fed into the decoder to produce multi-scale change maps $\{\text{M}_i \}^5_{i=1}$. These masks are combined to generate the final change map $\textbf{M}$. For supervised training, all predicted masks are bilinearly upsampled to match the input image resolution, enabling direct loss computation and network optimization.

\subsection{Spatial-Temporal Feature Extraction}

The siamese network extracts multi-scale spatiotemporal features through a weight-shared VGG16\_bn backbone, enhanced by a Feature Pyramid Network (FPN) and Channel Swap Module (CSM)‌. It processes image pairs $\textbf{X}$ and $\textbf{Y}$ to generate hierarchical features $ \{ \text{F}_\text{X}^{\text{CSM}_i}, \text{F}_\text{Y}^{\text{CSM}_i} \}^5_{i=1}$, where FPN improves scale adaptability through lateral cross-layer fusion, and CSM enables temporal interaction via channel-wise feature recalibration. The process can be expressed as:
\begin{equation}
\text{F}_\text{X}^{\text{CSM}_i}, \text{F}_\text{Y}^{\text{CSM}_i} = 
\begin{cases}
{CSM}(\text{F}_\text{X}^{\text{Conv}_i}, \text{F}_\text{Y}^{\text{Conv}_i}), \quad i=5 \\
 {CSM}(\text{F}_\text{X}^i, \text{F}_\text{Y}^i), \quad i = 4, 3, 2, 1
\end{cases}
\end{equation}
where ${CSM}(\cdot, \cdot)$ denotes the channel swap operation. The intermediate features $\text{F}_\text{X}^i$ and $\text{F}_\text{Y}^i$ are computed as follows:
\begin{equation}
\begin{cases}
\text{F}_\text{X}^i = \text{Conv}_2([\text{F}_\text{X}^{\text{Conv}_i}, {Up}(\text{F}_\text{X}^{\text{CSM}_{i+1}})], 3, 3) \\
\text{F}_\text{Y}^i = \text{Conv}_2([\text{F}_\text{Y}^{\text{Conv}_i}, {Up}(\text{F}_\text{Y}^{\text{CSM}_{i+1}})], 3, 3) 
\end{cases}
\end{equation}
where $\text{Conv}_2(\cdot, 3, 3)$ denotes the convolution layers with two $3 \times 3$ kernels, $Up(\cdot)$ refers to bilinear upsampling, and $[ \cdot, \cdot ]$ indicates feature concatenation along the channel dimension.

\subsubsection{Channel Swap Module (CSM)}
\begin{figure}[htb]
    \centering
    \includegraphics[width=0.95\columnwidth]{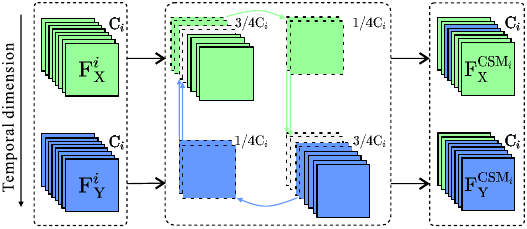}
    \caption{The architecture of CSM. It exchanges channel-level features between image pairs to model temporal dependencies.}
    \label{fig:csm}
\end{figure}

The CSM mitigates temporal discrepancies in change detection by dynamically exchanging channel-level features between dual-temporal image pairs. Significant foreground-background differences caused by time intervals often hinder accurate feature extraction and change region detection. Traditional siamese networks, lacking explicit temporal correction modeling, exhibit suboptimal sensitivity to genuine changes‌. To address this, CSM employs a channel-swapping operation (Figure~\ref{fig:csm}), which blends complementary temporal information while preserving dominant patterns.

\subsection{Change Feature Refinement}
The refinement stage progressively optimizes change representations through three core modules: Change-Guided Feature Fusion (CGFF), Feature Excitation and Suppression Module (FESM), and Pyramid-Aware Spatial-Channel Attention (PASCA). Specifically, the CGFF generates initial change features using deep-layer change priors for semantic guidance. The FESM dynamically reweights features to amplify change-sensitive patterns while reducing background interference, capturing both obvious and subtle changes. The PASCA employs multi-receptive fields and spatial-channel attention to focus on change regions across multiple scales.

\subsubsection{Change-Guided Feature Fusion (CGFF)}
\begin{figure}[!htb]
    \centering
    \includegraphics[width=0.95\columnwidth]{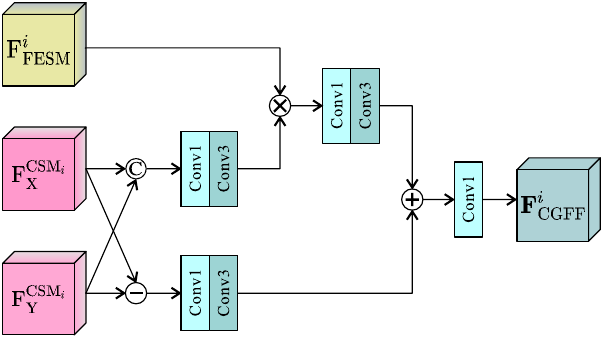}
    \caption{The architecture of CGFF. It adaptively fuses bi-temporal features guided by refined change features through bi-temporal concatenation, absolute difference computation, and change-guided feature fusion.}
    \label{fig:cgff}
\end{figure}

As shown in Figure~\ref{fig:cgff}, the CGFF module adaptively fuses bi-temporal features $\text{F}_\text{X}^{\text{CSM}_i}$ and $\text{F}_\text{Y}^{\text{CSM}_i}$ under the guidance of refined change features $\text{F}_{\text{FESM}}^i$. The process concatenates the bi-temporal features, computes their absolute difference, and use change-guided feature fusion to generate change features.
The formulations are as follows:
\begin{equation}
\text{F}_{\text{Con}} = \text{Conv}_2 ([\text{F}_\text{X}^{\text{CSM}}, \text{F}_\text{Y}^{\text{CSM}}], 1, 3)
\end{equation}
\begin{equation}
\text{F}_{\text{AD}} = \text{Conv}_2 (abs(\text{F}_\text{X}^{\text{CSM}} - \text{F}_\text{Y}^{\text{CSM}}), 1, 3) 
\end{equation}
% %
\begin{equation}
\text{F}_\text{CGFF}^i = 
\begin{cases}
\text{F}_\text{AD}^i, & i = 5 \\
\text{Conv}_1 (\text{F}_\text{Con}^{'i}+ \text{F}_\text{AD}^i,1), & i = 4,3,2,1
\end{cases}
\end{equation}
Here, $\text{F}_\text{Con}^{'}=\text{Conv}_2 (\text{F}_\text{Con} \odot \text{F}_\text{FESM},1,3)$, and $\odot$ indicates element-wise multiplication. $abs(\cdot)$ is the absolute operation, $\mathrm{Conv}_2(\cdot, 1, 3)$ denotes sequential convolution layers with kernel sizes of $1\times1$ and $3\times3$, and $\text{Conv}_1(\cdot, 1)$ represents a $1 \times 1$ convolution layer.

\subsubsection{Feature Excitation and Suppression Module (FESM)}
\begin{figure*}[!htb] % 使用figure*实现跨双栏排版
    \centering
    \includegraphics[width=0.85\textwidth]{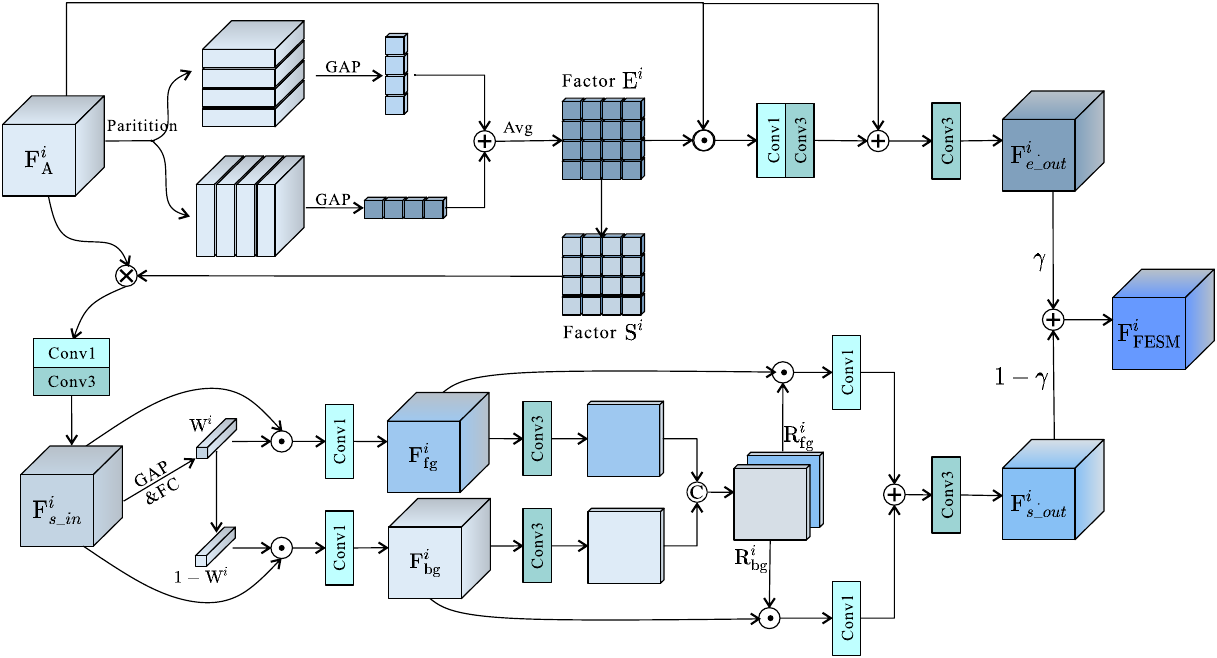} % 使用\textwidth作为宽度基准
    \caption{The architecture of FESM. It consists of two branches: the excitation branch enhances prominent change regions, while the suppression branch captures subtle changes by boosting foreground and suppressing background.}
    \label{fig:fesm}
\end{figure*}

Designed to amplify both prominent and subtle change patterns while suppressing noise interference, the core innovation lies in dual-axis adaptive feature recalibration through dynamic spatial-wise importance adjustment, significantly improving sensitivity to fine-grained changes. As shown in Figure~\ref{fig:fesm}, FESM operates by partitioning feature maps along dual-axis to compute:
\ding{182} Excitation branch: generates region-importance scores to enhance salient changes, producing activated features $\text{F}_{{e\_out}}$. 
\ding{183} Suppression branch: calculates region-suppression scores to obtain initial suppressed features $\text{F}_{{s\_in}}$, then refines them via foreground-boosting and background-suppression operations to generate $\text{F}_{{s\_out}}$.
The final output $\text{F}_{\text{FESM}}$ integrates the two branches using learnable fusion weights, as expressed by: 
\begin{equation}
\text{F}_{\text{FESM}} = \gamma \cdot \text{F}_{e\_out} + (1 - \gamma) \cdot \text{F}_{s\_out} , i=4,3,2,1
\end{equation}
where $\gamma$ is an adaptive coefficient. $\text{F}_{e\_out}$ and $\text{F}_{s\_out}$ are the output features of both branches.

\textbf{Excitation Branch.}
The branch adaptively strengthens salient change regions by applying multiple operations on features $\text{F}_\text{A}$, whose formulation varies across different layers:
\begin{equation}
\text{F}_\text{A}^i = \begin{cases} 
\text{F}_{\text{PASCA}}^{i+1}, & i = 4, 3 \\
{Up}(\text{Conv}_2(\hat{\text{M}}_c + \text{F}_\text{A}^{'i}, 1, 3)), & i = 2, 1
\end{cases} 
\end{equation}
Here, $\text{F}_\text{A}^{'i} = \text{Conv}_1(\hat{\text{M}}_c \odot \text{Conv}_1(\text{F}_{\text{PASCA}}^{i+1}, 3), 3)$, and $\hat{\text{M}}_c = \text{Conv}_1( \hat{\text{M}},1)$. $\text{Conv}_1(\cdot,3)$ denotes a \(3 \times 3\) convolution layer, and $\hat{\text{M}}$ is the initial change map.

Dual-Axis Partitioning: Feature $\text{F}_\text{A}$ is partitioned into $k$ sub-regions along both the horizontal and vertical axes. The region-importance scores ${E} \in R^{k \times k}$ are computed as:
\begin{equation}
\begin{cases}
{E}_{W} = {SF}({GAP}(\text{Conv}_1({S}_{W}(\text{F}_\text{A}, k), 1))) \\
{E}_{H} = {SF}({GAP}(\text{Conv}_1({S}_{H}(\text{F}_\text{A}, k), 1))) 
\end{cases}
\end{equation}
\begin{equation}
{E} = \frac{{E}_{W}+{E}_{H}}{2} 
\end{equation}
where ${S}_{W}(\cdot, k)$ and ${S}_{W}(\cdot, k)$ are horizontal and vertical splitting operations dividing $\text{F}_\text{A}$ into $k$ sub-regions. $GAP(\cdot)$ denotes global average pooling, and $SF(\cdot)$ represents Softmax normalization. $k$ is the partition factor. 

Feature Modulation: 
The importance matrix ${E}$ modulates feature responses to generate $\text{F}_{{e\_out}}$ by: 
\begin{equation}
\text{F}_{{e\_out}} = \text{Conv}_1\left({F}_{A} + \text{Conv}_2({E} \odot \text{F}_\text{A}, 1, 3), 3\right) .
\end{equation}

\textbf{Suppression Branch}
The suppression branch selectively suppresses the salient change regions using a suppression score ${S}$, resulting in a feature map $ F_{{s\_in}} $ that focuses on non-salient regions:
\begin{equation}
\text{F}_{{s\_in}} = \text{Conv}_2(\text{F}_\text{A} \odot {S}, 1, 3) 
\end{equation}
The region-suppression matrix $S \in R^{k \times k}$ is derived from $E$ to suppress high-score regions, highlighting regions with relatively lower scores. It is computed as:
\begin{equation}
s_{mn} = \begin{cases}
1 - \beta, & \text{if } e_{mn} = \max({E}) \\
1, & \text{otherwise}
\end{cases}
% , \quad \forall m,n = 1, 2, \dots, k 
\end{equation}
where $\beta \in [0,1] $ controls suppression intensity, and $e_{mn}$ is the score of the $(m, n)$-th sub-region in $E$. 

Foreground-Background Disentanglement:
The channel-wise attention weight ${W}$ is computed by:
\begin{equation}
{W} = {SG}(\text{Conv}_1({GAP}(\text{F}_{{s\_in}}), 1)) 
\end{equation}
where $SG(\cdot)$ is sigmoid function. The $\text{F}_{s\_in}$ are decomposed into foreground and background features $\text{F}_{fg}$ and $\text{F}_{bg}$ as:
\begin{equation}
\begin{cases}
\text{F}_{{fg}} = \text{F}_{{s\_in}} \odot W \\
\text{F}_{{bg}} = \text{F}_{{s\_in}} \odot (1-W) 
\end{cases}
\end{equation}
Feature Recalibration: The excitation factor $R_{fg}$ and background suppression factor $R_{bg}$ are obtained by:
\begin{equation}
{R}_{{fg}}, {R}_{{bg}} = {SF}\left([\text{Conv}_1(\text{F}_{{fg}}, 1), \text{Conv}_1(\text{F}_{{bg}}, 1)]\right)
\end{equation}
The suppression output $\text{F}_{\text{s\_out}}$ combines both paths:
\begin{equation}
\text{F}_{{s\_out}} = \text{Conv}_1\left( \text{F}_{{fg}} \odot {R}_{{fg}} + \text{F}_{{bg}} \odot {R}_{{bg}}, 3 \right)
\end{equation}

\begin{figure*}[!htb] % 跨双栏浮动环境
    \centering % 图片居中
    \includegraphics[width=0.85\textwidth]{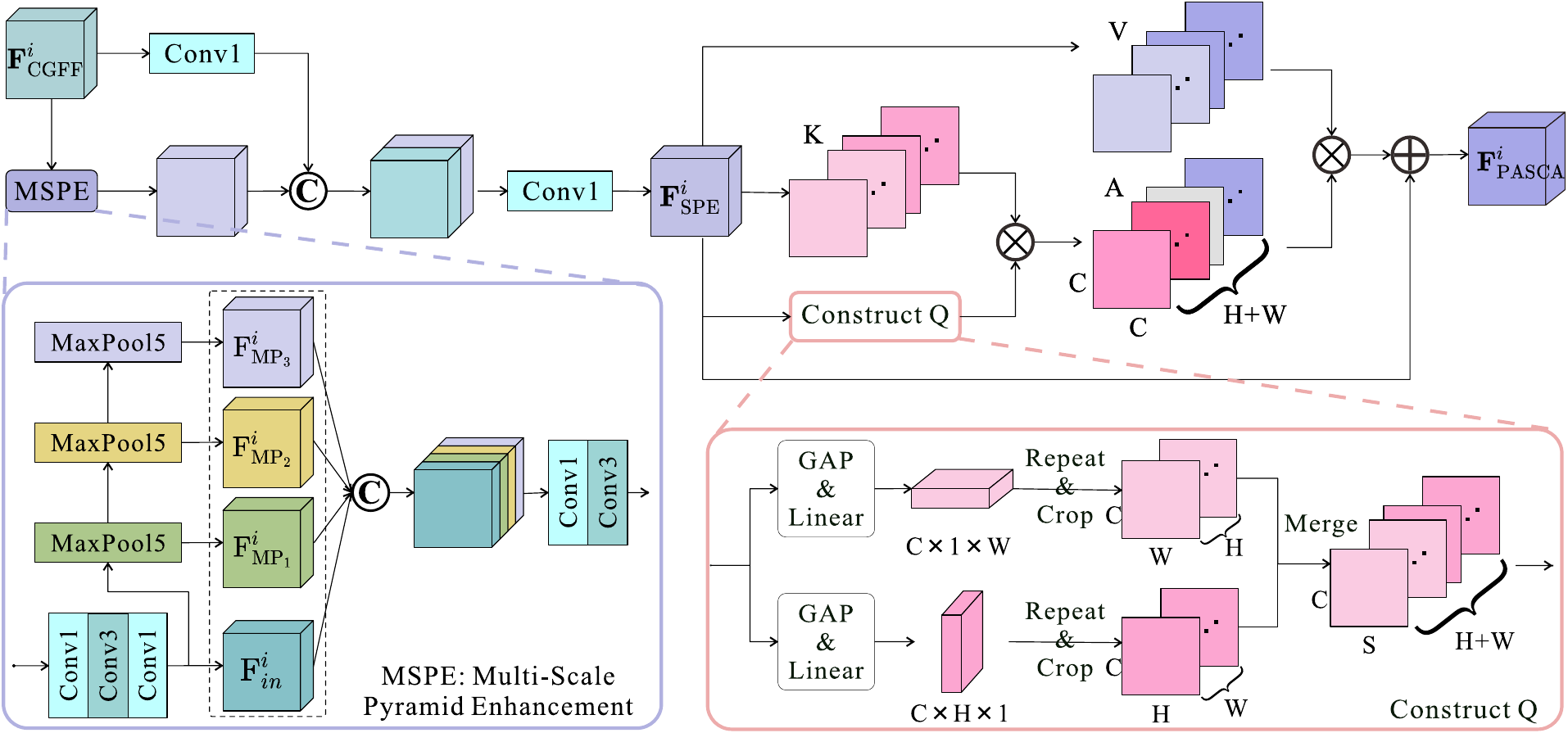}
    \caption{The architecture of PASCA. It enhances multi-scale change detection via pyramid-based feature fusion and spatial-channel attention, enabling localization of changes at various scales.}
    \label{fig:pasca}
\end{figure*}

\subsubsection{Pyramid-Aware Spatial-Channel Attention (PASCA)}

The PASCA effectively enhances the network's ability to perceive change regions across different scales by integrating multi-scale pyramid enhancement and spatial-channel attention mechanisms, as illustrated in Figure~\ref{fig:pasca}.

\textbf{Multi-Scale Pyramid Enhancement.} 
Convolutions and consecutive max-pooling operations are applied on $\text{F}_\text{CGFF}$ to extract multi-scale features, yielding $\text{F}_\text{MSPE}$, as follows:
\begin{equation}
\text{F}_{in}=\text{Conv}_3 (\text{F}_\text{CGFF},1,3,1)
\end{equation}
\begin{equation}
\text{F}_\text{MSPE}=\text{Conv}_2 ([\text{F}_{in}, \text{F}_{\text{MP}_1}, \text{F}_{\text{MP}_2}, \text{F}_{\text{MP}_3}],1,3)
\end{equation}
where $\mathrm{Conv}_3(\cdot, 1, 3, 1)$ denotes sequential convolutions with kernel sizes of $1\times1$, $3\times3$, and $1\times1$. $\text{F}_{\text{MP}_1}, \text{F}_{\text{MP}_2}, \text{F}_{\text{MP}_3}$ are the outputs of one, two and three max-pooling layers, respectively. 

To preserve the original information, a residual branch combines the original features with $\text{F}_\text{MSPE}$ to produce $\text{F}_\text{SPE}$:
\begin{equation}
\text{F}_\text{SPE}=\text{Conv}_1 ([ \text{F}_\text{MSPE}, \text{Conv}_1 (\text{F}_\text{CGFF},1)],1)
\end{equation}

\textbf{Spatial-Channel Attention (SCA).}
The SCA~\cite{song2022fully} is applied on $\text{F}_\text{SPE}$ to generate the refined $\text{F}_\text{PASCA}$, adaptively emphasizing crucial change regions. By jointly modeling spatial and channel information, it avoids overlooking small objects (from channel-only modeling) and disrupting large objects (from spatial-only modeling). The process is expressed as:
\begin{equation}
\text{F}_\text{PASCA} = \text{F}_\text{SPE} + \gamma \cdot( \sum_{m=1}^{C} \text{A}_{m} \cdot V ) 
\end{equation}
where $\gamma$ is a learnable parameter.    
The features $Q$, $K$, and $V$ are constructed from $\text{F}_\text{SPE}$ with $Q=K=V \in \mathbb{R}^{(H+W)\times C \times S}$, where $S=H=W$. 
The attention weight matrix $A$ is calculated as:
\begin{equation}
	A_{m,n} = \frac{exp(Q_m \cdot K^T_n)}{ {\textstyle \sum_{m=1}^{C}exp(Q_m \cdot K^T_n)} }, m,n \in {1,2,...,C} 
\end{equation}
where $A_{m,n}$ reflects channel-wise relative attention weights.
The construction process of \(Q\), \(K\), and \(V\) is designed to enrich feature representation by capturing information in both spatial and channel dimensions. Specifically, the input feature $\text{F}_\text{SPE}$ is segmented along the height ($H$) and width ($W$) dimensions, yielding two groups of slices: \(H\) slices of size \(C\times W\) and \(W\) slices of size \(C\times H\). Since \(H = W\), these slices are stacked and fused to obtain $Q=K=V \in \mathbb{R}^{(H+W)\times C \times S}$.

By integrating multi-scale aggregation and attention mechanisms, the PASCA effectively utilizes information at different scales and highlights change regions, ensuring robust and accurate change detection.

% Decoder block 
\begin{figure}[tb]
    \centering
    \includegraphics[width=0.6\linewidth]{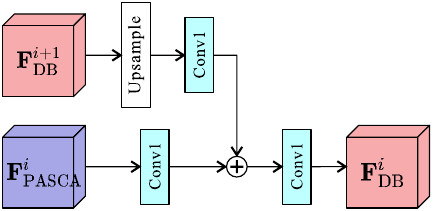}
    \caption{The architecture of DB. It propagates semantic-rich deep features to shallow layers via cross-scale fusion, integrating global context while retaining spatial details.}
    \label{fig:db}
\end{figure}

\subsection{Multi-Scale Change Detector}
The change detector hierarchically fuses refined change features across multiple scales to generate final change maps. As shown in Figure~\ref{fig:framework}, it uses five decoder blocks to progressively enhance features and predict change masks at different scales.

\subsubsection{Decoder Block}
Each decoder block propagates semantic-rich deep features to adjacent shallow layers through cross-scale fusion, leveraging global context while preserving spatial details. 
As illustrated in Figure~\ref{fig:db}, the high-level features $\text{F}_\text{DB}^{i+1}$ are unsampled and added to shallow features $\text{F}_\text{PASCA}^{i}$:
\begin{equation}
\text{F}_\text{DB}^{i}=
\begin{cases}
\mathrm{Conv}_{1}(\text{F}_\text{PASCA}^{5},1), & i = 5 \\
\mathrm{Conv}_{1}(\mathrm{Conv}_{1}({Up}(\text{F}_\text{DB}^{i + 1}),1) \\ 
\quad +\mathrm{Conv}_{1}(\text{F}_\text{PASCA}^{i},1),1), & i = 4,3,2,1
\end{cases}
\end{equation}

Multi-scale change masks $\{\text{M}_i\}^5_{i=1}$ are obtained by applying convolution operations on $\text{F}_\text{DB}^{i}$. The initial change map $\hat{\text{M}}$ is generated by concatenating and convolving features $\text{M}_{5}$, $\text{M}_{4}$ and $\text{M}_{3}$. The final change map $\text{M}$ is produced by integrating $\hat{\text{M}}$ with $\text{M}_{2}$ and $\text{M}_{1}$. The process is formulated as follows:
\begin{equation}
\text{M}_{i}= \text{Conv}_{2}(\text{F}_\text{DB}^{i},3,1),i=5,4,3,2,1
\end{equation}
\begin{equation}
\hat{\text{M}}=\text{Conv}_{1}([\text{M}_{5},\text{M}_{4},\text{M}_{3}],1)
\end{equation}
\begin{equation}
\text{M}=\text{Conv}_{1} ([\hat{\text{M}},\text{M}_{2},\text{M}_{1}],1 )
\end{equation}
The final change map $\text{M}$ can accurately identify changes.

\subsection{Loss Function}
Our method formulates change detection as a binary classification task, aiming to identify whether a target object has undergone changes while disregarding semantic categories. 
We employ the cross-entropy loss function to quantify the pixel-wise prediction errors:
\begin{equation}
    \mathcal{L}_{ce}=-([y\ln(p)+(1-y)\ln(1-p)]
\end{equation}
where $ y \in \{0, 1\} $ denotes the ground truth,  and \( p \) represents the  predicted probability.

To ensure multi-scale consistency and refine predictions progressively, the total loss integrates supervision from three sources: intermediate multi-scale masks, initial change map and final change map.
The total loss function is:
\begin{equation}
\begin{split}
\mathcal{L}_{\text{Total}} &= \mathcal{L}_{\text{ce}} (\hat{\text{M}}, \text{M}_{\text{GT}}) + \mathcal{L}_{\text{ce}} (\text{M}, \text{M}_{\text{GT}}) \\
&\quad + \sum_{i=1}^{5} \mathcal{L}_{\text{ce}} (\text{M}_{i}, \text{M}_{\text{GT}})
\end{split}
\end{equation}
Here, $\text{M}_\text{GT}$ is the ground-truth. All intermediate masks are bilinearly upsampled to the resolution of $\text{M}_\text{GT}$.

%%%%%%%%%%%%%%%%%%%%%%%%%%%%%%%%%%%%%%%%%%
\section{Experiments}
\label{experiment}

\subsection{Experimental Setups}

\subsubsection{Datasets}

\begin{table*}[htbp]
    \centering
    \centering{\caption{Dataset details.\label{tab:dataset}}}
    \resizebox{0.8\textwidth}{!}{
        \begin{tabular}{c | c c c c c}
            \toprule
            Dataset & Image pairs & Resolution & Train/Test & Change ratio &  Categories \\
            \midrule
            VL-CMU-CD & 1362 & $1024\times 768\times 3$ & $16005/332$ & 0.0716 & \makecell{bins, signs, vehicles, \\ maintenance, \textit{etc}.}  \\
            PCD & 200 & $1024\times 224\times 3$ & $8000/120$ & 0.2587 & \makecell{trees, buildings, \\ vehicles, \textit{etc}.} \\
            CDnet & 50209 & \makecell[c]{\(320\times 240 \times 3,\) \\ \(720\times 576 \times 3\) } & $40148/10061$ & 0.0479 & \makecell{person, vehicles, \\ boats, \textit{etc}. }  \\
            LEVIR-CD & 637 & $1024\times 1024\times 3$ & $18015/4674$ & 0.1116 & buildings  \\
            CDD & 11 & \makecell[c]{\(4725\times 2700 \times 3,\) \\ \(1900\times 1000 \times 3\) } & $12998/3000$ & 0.1234 & \makecell[c]{building, vehicles, \\ roads, \textit{etc}. } \\
            \bottomrule
        \end{tabular}
    }
\end{table*}
We conduct experiments on three common street-view CD datasets (VL-CMU-CD, PCD, and CDnet) and two remote sensing CD datasets (LEVIR-CD and CDD). Table.~\ref{tab:dataset} summarizes the details of these datasets.
\begin{itemize}
    \item ‌\textbf{VL-CMU-CD}‌~\cite{alcantarilla2018street}: Contains 1362 image pairs images in Pennsylvania, USA, from 152 real-world sequences with annotations. The resolution is \(1024 \times 768\) pixels.
    ‌\item \textbf{PCD}~\cite{jst2015change}: A panoramic CD dataset with ``TSUNAMI'' and ``GSV'' subsets. ``TSUNAMI'' has 100 pairs of pre- and post- Japanese tsunami panoramic images with change maps. ``GSV'' has 100 pairs of Google street-view panoramic images. The resolution is \(1024 \times 224\) pixels.
    ‌\item \textbf{CDnet‌}~\cite{wang2014cdnet}: Consists of nearly 50000 images from 31 video sequences, shot by various sensors. The resolution ranges from \(320 \times 240\) to \(720 \times 480\). It covers indoor and outdoor scenes and is divided into six challenge types.
    \item \textbf{LEVIR-CD}~\cite{chen2020spatial}: Includes 637 VHR image pairs of \(1024 \times 1024\) pixels from 20 USA regions. It covers various buildings and records over 31000 change instances.
    \item \textbf{CDD}‌~\cite{ji2018fully}: Contains VHR aerial images from 11 regions including New Zealand and multiple cities. The resolution ranges from \(1900 \times 1000\) to \(4725 \times 2700\) pixels.
\end{itemize}
In line with IDET‌~\cite{guo2025idet}, we partition the five datasets, perform data augmentation, and resize street-view images to \(320 \times 320\) pixels and remote sensing images to \(256 \times256\) pixels for training and testing.

\subsubsection{Baselines}
We compare our method with sixteen existing CD methods:
FCNCD~\cite{long2015fully}, ADCDnet~\cite{huang2020change}, CSCDnet~\cite{sakurada2020weakly}, IFN~\cite{zhang2020deeply}, BIT~\cite{chen2021remote}, STANet~\cite{chen2020spatial}, SNUNet~\cite{fang2021snunet}, ChangeFormer~\cite{bandara2022transformer}, TinyCD~\cite{codegoni2023tinycd}, Changer~\cite{fang2023changer}, ARCDNet~\cite{li2023towards}, SGSLN~\cite{zhao2023exchanging}, STENet~\cite{pan2024stenet}, DGMA$^{2}$-Net~\cite{ying2024dgma}, DDLNet~\cite{ma2024ddlnet}, and IDET~\cite{guo2025idet}.
Their descriptions are as follows:
\begin{itemize}
\item ‌FCNCD‌: First concatenates two images into a single six-channel image, then processes it through a fully convolutional network for change detection.
\item ‌ADCDnet‌: Uses a Siamese network for multi-scale feature extraction. Corresponding features are subtracted to obtain difference features, which are then fused hierarchically to generate change detection results.
\item ‌CSCDnet‌: Extracts multi-level features via a Siamese network, estimates optical flow between images using an optical flow estimation network to align features, and performs hierarchical fusion and alignment for change detection.
\item ‌IFN‌: Concatenates feature pairs from a Siamese network along the channel dimension, then predicts changes using a difference discrimination network similar to a fully convolutional network.
\item ‌BIT‌: Extracts image pair features via a CNN-based Siamese network, then employs Transformer encoders and decoders to learn semantic tokens for enhancing original features. The enhanced features are differenced to detect changes.
\item ‌STANet‌: Processes raw features using spatio-temporal attention modules to interactively leverage channel-wise semantic information and spatial positional information, then discriminates change maps via distance metrics.
\item ‌SNUNet‌: A densely connected Siamese network combining Siamese architecture and nested U-Net, integrating channel attention mechanisms to fuse and refine multi-level semantic features under deep supervision for more precise change detection.
\item ‌ChangeFormer‌: Adopts hierarchical Transformer encoders for feature extraction. The concatenated change features are fused through a multi-layer perceptron (MLP) to generate the final change map.
\item ‌TinyCD‌: A lightweight change detection network based on Siamese U-Net and EfficientNet~\cite{tan2019efficientnet}. It employs hybrid strategies for feature fusion to obtain difference features and uses MLPs for pixel-level classification.
\item ‌Changer‌: Processes feature alignment during feature extraction using interactive strategies like aggregated distributions and feature swapping, while introducing a flow-based dual alignment fusion module for effective alignment and feature fusion.
\item ARCDNet: Employs knowledge review modules to mine multi-level temporal differences, integrates online uncertainty estimation for uncertainty-aware learning, and fuses these features to generate change maps.
\item SGSLN: A binary change detection method with an exchanging dual encoder-decoder structure, integrating semantic guidance and spatial localization to address ‌bitemporal feature fusion at decision level‌ and ‌semantic-guided change area identification‌.
\item STENet: A dual encoder-decoder method integrating semantic-spatial cues to address ‌decision-layer fusion and ‌semantic-driven detection‌ in bitemporal change analysis.
\item DGMA$^2$-Net: A difference-guided multiscale aggregation network leveraging multiscale difference fusion, difference aggregation modules, and difference-attention refinement for change feature optimizaiton.
\item DDLNet: A dual-domain (frequency/spatial) RSCD network leveraging ‌frequency enhancement to amplify critical changes and spatial detail fusion to recover fine-grained spatiotemporal features, enabling robust change detection.
\item ‌IDET‌: Extracts features via a CNN-based Siamese network, iteratively enhances original and change features using a Transformer structure, and progressively fuses enhanced change features from coarse to fine for detection.
\end{itemize}
All methods are implemented with PyTorch and trained on the same datasets.

\subsubsection{Implementation Details}
The experiments are implemented with PyTorch and trained on a single NVIDIA GeForce 2080Ti GPU. We adopt the Adam optimizer with an initial learning rate of 1e-3, a momentum of 0.9, and a weight decay of 0.999. The batch size is configured to 4. All methods are trained for 20 epochs on three common street-view CD datasets and 200 on two remote sensing CD datasets.

\subsubsection{Evaluation Metrics}
To accurately evaluate the performance of different methods, we adopt five commonly used evaluation metrics: Precision ($P$), Recall ($R$), F1-measure ($F1$), Overall Accuracy ($OA$), and Intersection over Union ($IoU$). 
The formulas are as follows:
\begin{align}
    P&=\frac{TP}{TP + FP} \\
    R&=\frac{TP}{TP + FN} \\
    F1&=\frac{2\times P\times R}{P + R} \\
    OA&=\frac{TP + TN}{TP + TN+FP + FN} \\
    IoU&=\frac{TP}{TP + FP+FN}
\end{align}
where $TP$, $TN$, $FP$, and $FN$ denote true positive, true negative, false positive, and false negative, respectively. Higher $F1$ and $IoU$ values indicate better-performing methods, as they reflect more accurate positive-sample identification and better overlap between predicted and actual results.

\subsection{Main Results}

\subsubsection{Results on Common Street-view CD Datasets}
% main results on street-view CD datasets
\begin{table*}[!htbp]
    \centering
    \caption{Quantitative comparison of different CD methods on VL-CMU-CD, PCD and CDnet datasets. Note that \textcolor{rank1}{red} marks the best-performing method, \textcolor{rank2}{blue} the second-best, and \textcolor{rank3}{green} the third-best.}
    \label{tab:streetviewResults}
    \setlength{\tabcolsep}{6pt}
    \resizebox{0.98\textwidth}{!}{
    \renewcommand{\arraystretch}{1.1}
        \begin{tabular}{l|ccccc|ccccc|ccccc}
            \toprule
            \multirow{2}{*}{Method} & \multicolumn{5}{c|}{VL-CMU-CD} & \multicolumn{5}{c|}{PCD} & \multicolumn{5}{c}{CDnet}  \\
            & $P$ & $R$ & $F1$ & $OA$ & $IoU$ & $P$ & $R$ & $F1$ & $OA$ & $IoU$ & $P$ & $R$ & $F1$ & $OA$ & $IoU$  \\
            \midrule
            FCNCD & 84.2 & 84.3 & 84.2 & 98.4 & 72.9 
            & 68.7 & 61.7 & 65.0 & \textcolor{rank1}{94.1} & 48.0
            & 82.9 & 88.3 & 85.5 & 99.2 & 76.0 \\
            ADCDnet & 92.8 & 94.3 & \textcolor{rank3}{93.5} & \textcolor{rank2}{99.3} & \textcolor{rank3}{87.9} & \textcolor{rank2}{78.8} & 73.4 & \textcolor{rank2}{76.0} & \textcolor{rank2}{90.0} & \textcolor{rank2}{62.3} & 89.0 & 85.5 & 87.2 & 99.3 & 77.9 \\
            CSCDnet & 89.2 & 91.1 & 90.1 & 98.7 & 82.1 & 66.0 & 72.6 & 69.1 & 83.7 & 51.9 & \textcolor{rank3}{93.9} & 82.3 & 87.7 & 99.0 & 78.4 \\
            IFN & 93.1 & 75.9 & 83.6 & 98.1 & 71.4 & 69.5 & 75.7 & 72.5 & 87.2 & 57.7 & 93.1 & 80.0 & 86.1 & 98.9 & 75.3 \\
            BIT & 90.5 & 87.6 & 89.0 & 98.8 & 80.6 & \textcolor{rank3}{77.7} & 57.3 & 66.0 & 85.4 & 50.1 & \textcolor{rank2}{95.4} & 81.9 & 88.1 & 99.1 & 78.9 \\
            STANet & 73.1 & \textcolor{rank2}{95.8} & 82.9 & 98.0 & 70.7 & 69.6 & 52.7 & 60.0 & 79.9 & 41.2 & 72.2 & \textcolor{rank3}{94.3} & 81.8 & 98.3 & 69.2 \\
            SNUNet & 78.4 & 83.1 & 80.7 & 97.9 & 67.7 & 73.7 & 67.5 & 70.5 & 86.9 & 54.8 & 91.8 & 88.4 & 90.1 & \textcolor{rank3}{99.4} & 82.0 \\
            ChangeFormer & 88.0 & 88.6 & 88.3 & 98.7 & 79.2 & 74.0 & 69.2 & 71.5 & 86.6 & 56.4 & 93.4 & 85.9 & 89.5 & 99.2 & 81.3 \\
            TinyCD & 89.4 & 94.2 & 91.8 & 99.1 & 85.0 & 67.6 & \textcolor{rank1}{83.2} & \textcolor{rank3}{74.6} & 85.7 & 58.9 & 90.0 & 93.8 & 91.9 & 99.5 & 84.9 \\
            Changer & \textcolor{rank2}{93.1} & 90.7 & 91.9 & \textcolor{rank3}{99.2} & 85.1 & 73.4 & 64.2 & 68.5 & 86.5 & 52.7 & \textcolor{rank1}{95.7} & 89.7 & \textcolor{rank2}{92.6} & \textcolor{rank2}{99.6} & \textcolor{rank2}{87.2} \\
            ARCDNet & 76.3 & \textcolor{rank1}{97.5} & 85.6 & 98.4 & 74.8 & 67.2 & \textcolor{rank3}{78.9} & 72.3 & 89.2 & 60.8 & 74.7 & \textcolor{rank2}{94.4} & 83.4 & 98.4 & 71.8 \\
            SGSLN & 74.0 & \textcolor{rank1}{97.5} & 84.2 & 98.2 & 72.7 & 68.3 & \textcolor{rank3}{80.7} & 74.0 & 87.8 & \textcolor{rank3}{62.1} & 75.8 & 93.0 & 83.5 & 98.5 & 72.0 \\
            STENet & 89.1 & 80.5 & 84.6 & 98.4 & 73.6 & 72.0 & 74.5 & 73.2 & 87.3 & 57.8 & 77.0 & 88.6 & 82.4 & 98.4 & 70.6 \\
            DGMA$^2$-Net & 89.2 & 92.9 & 91.0 & 99.1 & 83.8 & 68.5 & 76.1 & 72.1 & 86.3 & 56.4 & 77.6 & 92.4 & 84.3 & 98.6 & 73.3 \\
            DDLNet & 88.6 & 94.8 & 91.6 & \textcolor{rank3}{99.2} & 84.5 & 77.0 & 68.5 & 72.5 & 88.9 & 57.9 & 85.4 & 90.7 & 88.0 & 99.0 & 79.0 \\
            IDET & \textcolor{rank1}{93.5} & 94.5 & \textcolor{rank2}{94.0} & \textcolor{rank1}{99.4} & \textcolor{rank2}{88.7} & 74.2 & 77.9 & \textcolor{rank2}{76.0} & 88.8 & 61.9 & 90.8 & 94.1 & \textcolor{rank3}{92.4} & \textcolor{rank1}{99.7} & \textcolor{rank3}{86.2} \\
            \rowcolor{gray!15}
            CEBSNet (\textbf{Ours}) & \textcolor{rank3}{93.0} & \textcolor{rank3}{95.5} & \textcolor{rank1}{94.2} & \textcolor{rank1}{99.4} & \textcolor{rank1}{89.1} & \textcolor{rank1}{79.2} & 74.8 & \textcolor{rank1}{76.9} & \textcolor{rank3}{90.1} & \textcolor{rank1}{63.0} & 93.1 & \textcolor{rank1}{95.5} & \textcolor{rank1}{94.3} & \textcolor{rank1}{99.7} & \textcolor{rank1}{89.1} \\
            \bottomrule
        \end{tabular}}
\end{table*}

\begin{figure*}[!htb]
    \centering
    \includegraphics[width=1\textwidth]{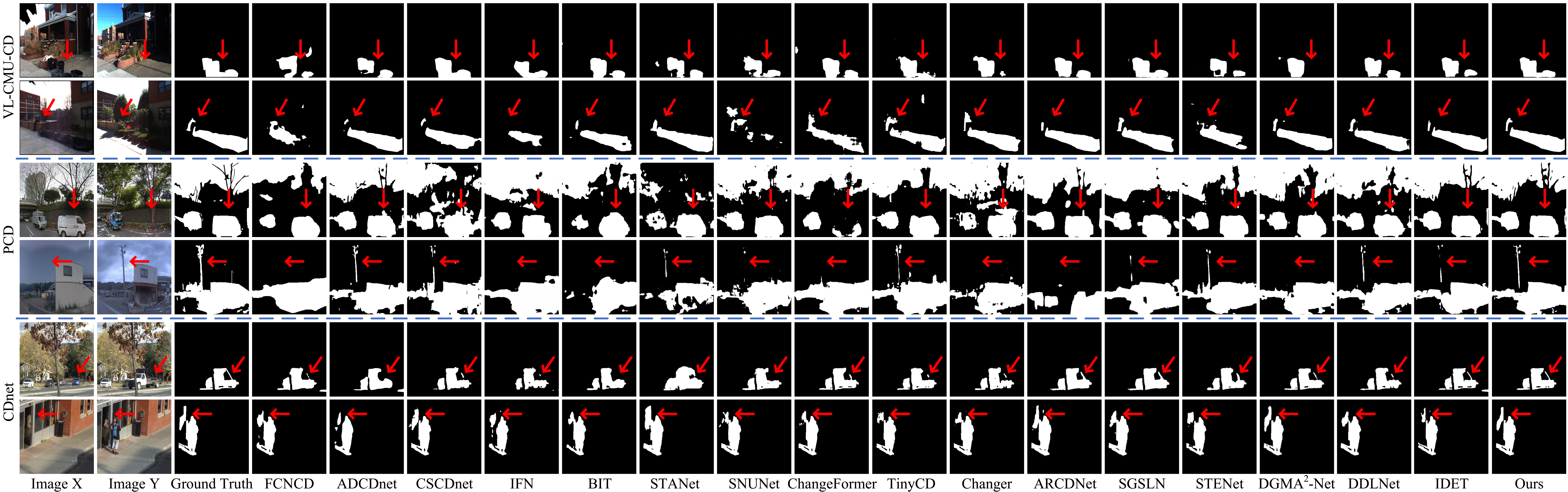}
    \caption{Detection results of different methods on VL-CMU-CD, PCD, and CDnet datasets. The \textcolor{red}{red arrows} indicate noticeable changes.}
    \label{fig:streetviewVis}
\end{figure*}

Table~\ref{tab:streetviewResults} presents the experimental results of our method and other comparative methods on three common street-view datasets, namely VL-CMU-CD, PCD, and CDnet. These results clearly show that our CEBSNet achieves the best performance across all three datasets, validating its generalizability in complex street-view scenarios. On VL-CMU-CD, our CEBSNet achieves the peak performance in $F1$, $OA$, and $IoU$ metrics, reaching \(94.2\%\), \(99.4\%\), and \(89.1\%\) respectively. When compared with the previous SOTA method IDET, the $F1$ and $IoU$ values are improved by \(0.2\%\) and \(0.4\%\) respectively. For PCD dataset, which contains highly challenging scenarios with diverse changes (e.g., buildings, trees, grasslands, vehicles), our method attains the highest $P$ (79.2\%), $F1$ (76.9\%), and $IoU$ (63.0\%). Notably, its $F1$ score is \(0.9\%\) higher than ADCDnet and IDET, indicating its effectiveness in handling complex changes. On CDnet, our CEBSNet excels in four metrics: $R$ (95.5\%), $F1$ (94.3\%), $OA$ (99.7\%), and $IoU$ (89.1\%). Compared with Changer, these metrics are increased by \(5.8\%\), \(1.7\%\), \(0.1\%\), and \(1.9\%\) respectively, further validating its superiority. These ‌results firmly prove that our CEBSNet excels in ‌common street-view change detection, showing its great potential for real-world applications.

Fig.~\ref{fig:streetviewVis} shows some typical examples of different methods on three common street-view datasets. Accurate and complete detection is crucial for identifying main change areas and handling details and noise. Most methods can detect main change regions, but struggle with details and background noise. Methods like FCNCD, IFN, Changer, STENet, and DDLNet tend to lose details when capturing main changes, thus affecting the judgement of changes. CSCDnet and BIT are easily interfered by the background noise, causing detection errors and inaccurate marking of change areas. STANet, SNUNet, and ChangeFormer are severely affected by background noise, sometimes even missing main change objects, resulting in incomplete and unreliable results. Among these methods, ADCDnet and IDET perform relatively better. But due to background noise from large time spans, they ignore the subtle changes.
Our method has distinct advantages. It can accurately detect change regions, keep rich details, and cover both significant and subtle changes, like large-scale building renovation or  subtle road-sign alterations. This shows its higher accuracy and stronger robustness, further validating its superior performance in common street-view change detection.

% main results on remote sensing CD datasets
\begin{table*}[htbp]
    \centering % 表格整体居中（跨双栏）
    \caption{Quantitative comparison of different CD methods on LEVIR-CD and CDD datasets. Note that \textcolor{rank1}{red} marks the best-performing method, \textcolor{rank2}{blue} the second-best, and \textcolor{rank3}{green} the third-best.}
    \label{tab:remotesensingResults}
    \resizebox{0.7\textwidth}{!}{ % 自动缩放至双栏宽度（关键修改）
        \begin{tabular}{l|ccccc|ccccc}
            \toprule
            \multirow{2}{*}{Method} & \multicolumn{5}{c|}{LEVIR-CD} & \multicolumn{5}{c}{CDD} \\
            & $P$ & $R$ & $F1$ & $OA$ & $IoU$ & $P$ & $R$ & $F1$ & $OA$ & $IoU$ \\
            \midrule
            FCNCD & 89.7 & 84.0 & 86.8 & 97.7 & 77.1 & 85.1 & 69.8 & 76.7 & 97.5 & 63.7 \\
            ADCDnet & 90.4 & 81.8 & 85.9 & 97.4 & 75.4 & 92.5 & 78.8 & 85.1 & 98.5 & 75.1 \\
            CSCDnet & 84.6 & 81.6 & 83.1 & 96.8 & 71.5 & 79.1 & 48.9 & 60.4 & 90.9 & 42.7 \\
            IFN & 89.7 & 79.8 & 84.5 & 96.9 & 72.7 & 94.2 & 69.2 & 79.8 & 97.7 & 66.5 \\
            BIT & \textcolor{rank3}{92.1} & 81.6 & 86.5 & 97.7 & 76.3 & 93.5 & 80.2 & 86.3 & 98.7 & \textcolor{rank3}{76.6} \\
            STANet & 84.8 & 69.9 & 76.6 & 95.3 & 61.8 & 71.6 & 69.6 & 70.6 & 95.0 & 54.6 \\
            SNUNet & 88.2 & 83.8 & 86.0 & 97.7 & 76.0 & 86.3 & 64.0 & 73.5 & 97.0 & 58.8 \\
            ChangeFormer & 90.7 & 85.5 & 88.0 & \textcolor{rank3}{98.0} & 79.0 & 89.2 & 73.0 & 80.3 & 98.0 & 68.0 \\
            TinyCD & 89.2 & 83.4 & 86.2 & 97.2 & 76.3 & \textcolor{rank2}{95.1} & 80.1 & \textcolor{rank2}{87.0} & 98.8 & \textcolor{rank3}{77.6} \\
            Changer & \textcolor{rank1}{92.9} & 74.8 & 82.9 & 97.0 & 70.4 & \textcolor{rank3}{94.8} & 73.0 & 82.5 & 98.4 & 70.9 \\
            ARCDNet & 91.4 & 86.5 & \textcolor{rank3}{88.9} & 97.9 & 80.0 & 81.8 & \textcolor{rank3}{83.4} & 82.6 & 97.3 & 71.2 \\
            SGSLN & 90.8 & \textcolor{rank1}{87.5} & \textcolor{rank2}{89.2} & \textcolor{rank2}{98.1} & \textcolor{rank2}{80.8} & 83.6 & \textcolor{rank2}{86.7} & 85.1 & 97.8 & 74.9 \\
            STENet & 91.7 & 70.4 & 79.7 & 95.6 & 66.4 & 68.5 & 53.0 & 59.7 & 93.9 & 43.0 \\
            DGMA$^2$-Net & 90.9 & 70.1 & 79.1 & 95.5 & 65.8 & 73.6 & 77.1 & 75.3 & 97.0 & 61.7 \\
            DDLNet & 90.9 & 84.4 & 87.5 & 97.7 & 77.8 & 91.6 & 80.2 & 85.5 & 98.5 & 75.3 \\
            IDET & 91.3 & \textcolor{rank3}{86.6} & \textcolor{rank3}{88.9} & \textcolor{rank2}{98.1} & \textcolor{rank3}{80.2} & 91.1 & 83.1 & \textcolor{rank3}{86.9} & 98.7 & \textcolor{rank2}{77.6} \\
            \rowcolor{gray!15}
            CEBSNet (\textbf{Ours}) & \textcolor{rank2}{92.7} & \textcolor{rank2}{87.2} & \textcolor{rank1}{89.9} & \textcolor{rank1}{98.2} & \textcolor{rank1}{81.8} & \textcolor{rank1}{99.1} & \textcolor{rank1}{87.1} & \textcolor{rank1}{92.7} & \textcolor{rank1}{99.3} & \textcolor{rank1}{86.5} \\
            \bottomrule
        \end{tabular}
    }
\end{table*}

\begin{figure*}[!htb]
    \centering
    \includegraphics[width=1\textwidth]{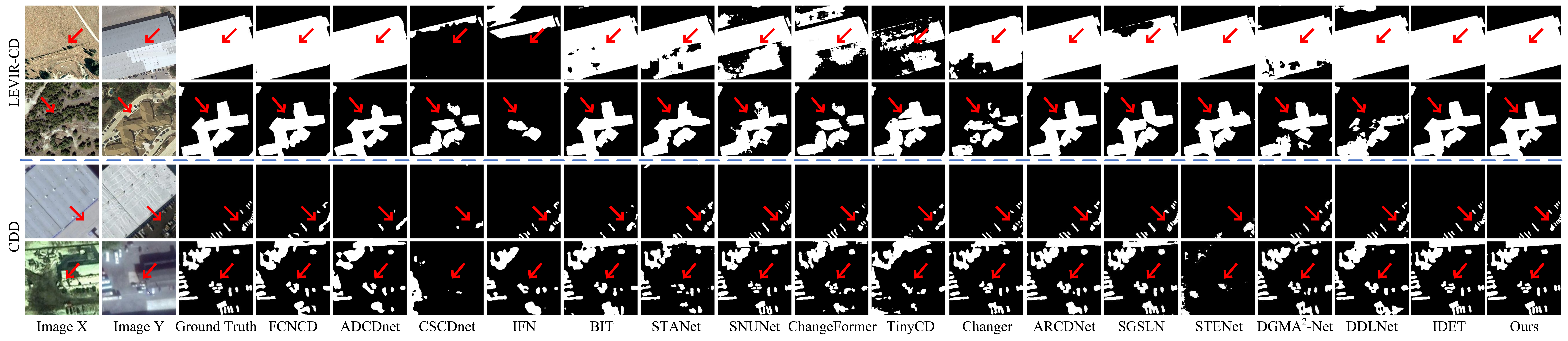}
    \caption{Detection results of different methods on LEVIR-CD and CDD datasets. The \textcolor{red}{red arrows} indicate noticeable changes.}
    \label{fig:remotesensingVis}
\end{figure*}

\subsubsection{Results on Remote Sensing CD Datasets}

Table~\ref{tab:remotesensingResults} presents the experimental results of our method and other comparative methods on two remote sensing datasets, namely LEVIR-CD and CDD. These results clearly demonstrate that our method achieves the optimal performance on both datasets, confirming its effectiveness in complex remote sensing scenarios. On LEVIR-CD, our CEBSNet attains the best results in $F1$, $OA$, and $IoU$ metrics, reaching \(89.9\%\), \(98.2\%\), and \(81.8\%\) respectively. Compared with the previously best-performing SGSLN, our method shows improvements of \(0.7\%\), \(0.1\%\), and \(1.0\%\) in these metrics. On CDD, our CEBSNet outperforms other methods in all five metrics of $R$, $P$, $F1$, $OA$, and $IoU$, with specific values of \(99.1\%\), \(87.1\%\), \(92.7\%\), \(99.3\%\), and \(86.5\%\) respectively. Notably, CEBSNet is the only method with an $F1$ score exceeding \(90.0\%\). When compared with TinyCD, our method shows significant improvements in all these five metrics, with the increments being \(4.3\%\), \(7.0\%\), \(5.7\%\), \(0.5\%\), and \(8.9\%\) respectively. These findings clearly show our method's outstanding performance and its great potential in remote sensing change detection.

Fig.~\ref{fig:remotesensingVis} shows some typical examples of different methods on two remote sensing datasets. In the 1st and 2nd rows, methods like ADCDnet, ARCDNer, SGSLN, IDET, and our CEBSNet can comprehensively detect building changes. Conversely, the remaining methods encounter issues like non-detection or partial detection failures, failing to accurately represent building change information. In the 3rd and 4th rows, ARCDNet, SGSLN, DGMA$^2$-Net, and our CEBSNet can precisely capture detailed information. when confronted with minuscule and dispersed change regions, most methods are highly susceptible to noise interference. This lead to miss minor changes, causing false positives and false negatives. Among all these methods, our CEBSNet showcases remarkable advantages in both accuracy and integrity. It can accurately identify main change regions and clearly reveal fine-grained details. Even for subtle change areas, it still yield satisfactory detection results. The visual evidence aligns with quantitative metrics, confirming the effectiveness and robustness of our CEBSNet in complex remote sensing change detection.

\subsubsection{Computational Complexity Analysis}
Table~\ref{tab:computationalCost} shows the computational costs and parameter counts across different methods on common street-view datasets with an image size of \(320 \times 320\). By conducting an in-depth comparison of the performance and complexity of each method, our method achieves the best performance across all datasets while maintaining appropriate costs in GFLOPs and Params. In terms of GFLOPs, our method exhibits lower computational complexity than methods like ADCDnet and IDET. Regarding Params, it requires fewer parameters than networks such as FCNCD, CSCDnet, and IDET. In conclusion, our method strikes a good balance between complexity and performance, enabling excellent detection performance with moderate complexity.
% computational costs and parameter
\begin{table}[h!]
\centering
\caption{Comparison of computational costs and parameter counts on common street-view datasets.}
\label{tab:computationalCost}
% \resizebox{0.36\textwidth}{!}{
\begin{tabular}{l|cc}
\toprule
Method & GFLOPs & Params (M) \\
\midrule
FCNCD &	100.2 & 134.2 \\
ADCDnet & 217.6 & 32.2 \\
CSCDnet	 & 65.9 & 92.3 \\
IFN	 & 128.5 & 50.4 \\
BIT	 & 53.8 & 3.4 \\
STANet	 & 20.1 & 16.8 \\
SNUNet	 & 85.7 &	12.0 \\
ChangeFormer & 316.9 & 29.7 \\
TinyCD & 2.4 & 0.27 \\
Changer	 & 9.1 & 11.3 \\
ARCDNet	 & 22.1 & 14.3 \\
SGSLN	 & 18.0 & 6.0 \\
STENet	 & 28.6 & 37.1 \\
DGMA$^2$-Net & 23.0 & 11.0 \\
DDLNet	 &	11.3 & 12.7 \\
IDET	 &	195.9 & 100.7 \\
\rowcolor{gray!15}
CEBSNet (\textbf{Ours})	 &	174.7	 &	86.0\\
\bottomrule
\end{tabular}
% }
\end{table}

\subsection{Ablation Studies}
In this section, we conduct three ablation experiments and one feature visualization analysis to thoroughly evaluate the key components and performance of our method. We begin by exploring the main components' impact on performance. Then, we analyze the optimal feature shift ratio in Channel Swap Module (CSM) for better temporal correlation modeling. After that, we study the $k$-value setting in Feature Excitation and Suppression Module (FESM). Finally, a feature visualization analysis is performed to shows each module's response during feature processing. Note that all ablation studies are conducted on VL-CMU-CD dataset.

\subsubsection{Component Effectiveness Analysis}
To thoroughly evaluate the contribution of each component in our method, we conduct a series of ablation experiments using a module-combination strategy. The experimental results on VL-CMU-CD dataset are presented in Table~\ref{tab:ablation_modules}, clearly showing the performance gains with different module combinations. The CSM enhances the network's noise resilience by modeling the temporal dependencies between image pairs. The CGFF enriches difference information with global context. The FESM improves the detection accuracy of change regions through excitation and suppression operations. The PASCA employs its multi-receptive fields and attention mechanism, contributing to more accurate change detection. In summary, the ablation results prove the rational design and effectiveness of each module in enhancing network performance, highlighting their importance.

% Ablation study
\begin{table}[h]
    \centering
    \caption{Ablation study with varying components on VL-CMU-CD dataset. Best metrics are highlighted in \textbf{bold}.}
    \label{tab:ablation_modules}
    \footnotesize  
    \setlength{\tabcolsep}{3pt}
    \resizebox{0.5\textwidth}{!}{ % 自动缩放至双栏宽度（关键修改）
    \begin{tabular}{@{} ccccc|ccccc @{}}  
        \toprule   
        {Base} & {CSM} & {CGFF} & {FESM} & {PASCA} &
        $P$ & $R$ & $F1$ & $OA$ & $IoU$ \\
        \midrule  
        $\checkmark$ & & & & & 91.36 & 94.63 & 92.96 & 99.23 & 86.85  \\  
        $\checkmark$ & $\checkmark$ & & & & 91.88 & 95.78 & {93.79} & 99.33 & 88.35  \\  
        $\checkmark$ & $\checkmark$ & $\checkmark$ & & & 92.51 & 95.34 & 93.90 & 99.35 & 88.58  \\  
        $\checkmark$ & $\checkmark$ & $\checkmark$ & $\checkmark$ & & 92.88 & 95.29 & 94.07 & \textbf{99.36} & 88.48  \\  
        \rowcolor{gray!15}
        $\checkmark$ & $\checkmark$ & $\checkmark$ & $\checkmark$ & $\checkmark$ & \textbf{92.96} & 95.47 & \textbf{94.20} & \textbf{99.36} & \textbf{89.08}  \\  
        \bottomrule
    \end{tabular}}
    \vspace{0.2cm}
    \begin{minipage}{\columnwidth}
        \raggedright
        \footnotesize
        \textbf{Note:} Components marked with $\checkmark$ are activated. Abbreviations: CSM (Channel Swap Module), CGFF (Change-Guided Feature Fusion), FESM (Feature Excitation and Suppression Module), PASCA (Pyramid-Aware Spatial-Channel Attention).
    \end{minipage}
\end{table}

\subsubsection{Impact of Channel Shift Ratio in CSM}
Table~\ref{tab:csm_swap} evaluates the channel shift ratio in Channel Swap Module (CSM), quantifying its impact. Our experiments with swap ratios $\{0, 1/4, 2/4, 3/4, 4/4\}$ reveal that a ‌1/4 ratio‌ achieves peak performance on VL-CMU-CD, motivating its adoption as the default configuration. The experiments indicate that moderate feature interaction enhances temporal correlation modeling and noise suppression, whereas excessive swapping degrades dependency learning. The 1/4 balance preserves discriminative spatio-temporal patterns while mitigating redundancy.
% CSM 
\begin{table}[H]  
\centering  
\caption{Ablation Analysis of feature shift ratio in CSM. Best scores in \textbf{bold}.}  
\label{tab:csm_swap}  
\begin{tabular}{c|ccccc}  
\toprule  
{Ratio} & ${P}$ & ${R}$ & ${F1}$ & ${OA}$ & ${IoU}$ \\  
\midrule  
$0$ & 91.32 & 95.38 & 93.31 & 99.29 & 87.53 \\  
\rowcolor{gray!15}
$1/4$ & {92.96} & 95.47 & \textbf{94.20} & \textbf{99.36} & \textbf{89.08} \\  
$2/4$ & 92.67 & \textbf{95.68} & 94.15 & \textbf{99.36} & 88.99 \\  
$3/4$  & 92.20 & 95.92 & 94.03 & 99.35 & 88.73 \\  
$4/4$  & \textbf{93.06} & 94.90 & 93.97 & 99.37 & 88.65 \\ 
\bottomrule  
\end{tabular}    
\end{table}

subsubsection{Impact of Hierarchical $k$-value Configuration in FESM}
Table~\ref{tab:fesm_k} details the setting of the $k$-value in Feature Excitation and Suppression Module (FESM). To explore its impact on model performance, we conduct two controlled experiments:
\begin{itemize} 
  \item {Dynamic-$k$ Group}: Fix pixel span to 4 while assigning layer-specific $k$-values (network levels $i=4,3,2,1$).  
  \item {Fixed-$k$ Group}: Maintain identical $k$-values across layers with adaptive pixel spans.  
\end{itemize}

% FESM
\begin{table}[htbp]  % 替换 [H] 为浮动选项，避免强制位置导致的溢出
    \centering  % 表格在单栏内居中
    \caption{Ablation analysis on $k$ configurations in FESM. Best scores in \textbf{bold}.}
    \label{tab:fesm_k}
    \footnotesize  % 小字体（可改为 \scriptsize 进一步压缩）
    
    % 压缩列间距并去除左右边距（关键优化）
    \setlength{\tabcolsep}{4pt}  % 列间距从默认6pt压缩至4pt
    \begin{tabular}{@{} c|c|ccccc @{}}  % @{} 去除表格左右边距，紧凑显示
        \toprule  
        {Type} & $\{k_4,k_3,k_2,k_1\}$ & ${P}$ & ${R}$ & ${F1}$ & ${OA}$ & ${IoU}$ \\ 
        \midrule  
        \multirow{3}{*}{Dynamic} & \{2,4,8,16\} & 92.98 & 95.02 & 93.99 & 99.36 & 88.69 \\  
        & \{4,8,16,32\} & 92.54 & 95.49 & 93.99 & \textbf{99.38} & 88.70 \\  
        \rowcolor{gray!15} & \{5,10,20,40\} & \textbf{92.96} & 95.47 & \textbf{94.20} & 99.36 & \textbf{89.08} \\  
        \midrule  
        \multirow{3}{*}{Fixed} & \{4,4,4,4\} & 92.57 & 95.48 & 94.00 & 99.37 & 88.71 \\  
        & \{5,5,5,5\} & 92.31 & \textbf{95.50} & 93.88 & 99.36 & 88.49 \\  
        & \{10,10,10,10\} & 92.86 & 95.37 & 94.10 & 99.36 & 88.91 \\   
        \bottomrule  
    \end{tabular}
\end{table}

On VL-CMU-CD, the {dynamic-$k$ configuration} with values $\{5,10,20,40\}$ achieves peak performance with \(94.20\%\) $F1$ and \(89.08\%\) $IoU$, outperforming fixed-$k$ baselines by \(0.21\%\) - \(0.59\%\) $F1$ points. This verifies that progressively increasing $k$-values in deeper layers (from $i=4$ to $i=1$) enhance multi-scale feature discrimination, whereas uniform $k$-values cause either oversuppression (small $k$) or noisy activation (large $k$). We therefore adopt $\{5,10,20,40\}$ as default configuration.

\begin{figure*}[!htb] % 跨双栏浮动环境（星号为关键标识）
    \centering % 图片居中
    \includegraphics[width=0.8\textwidth]{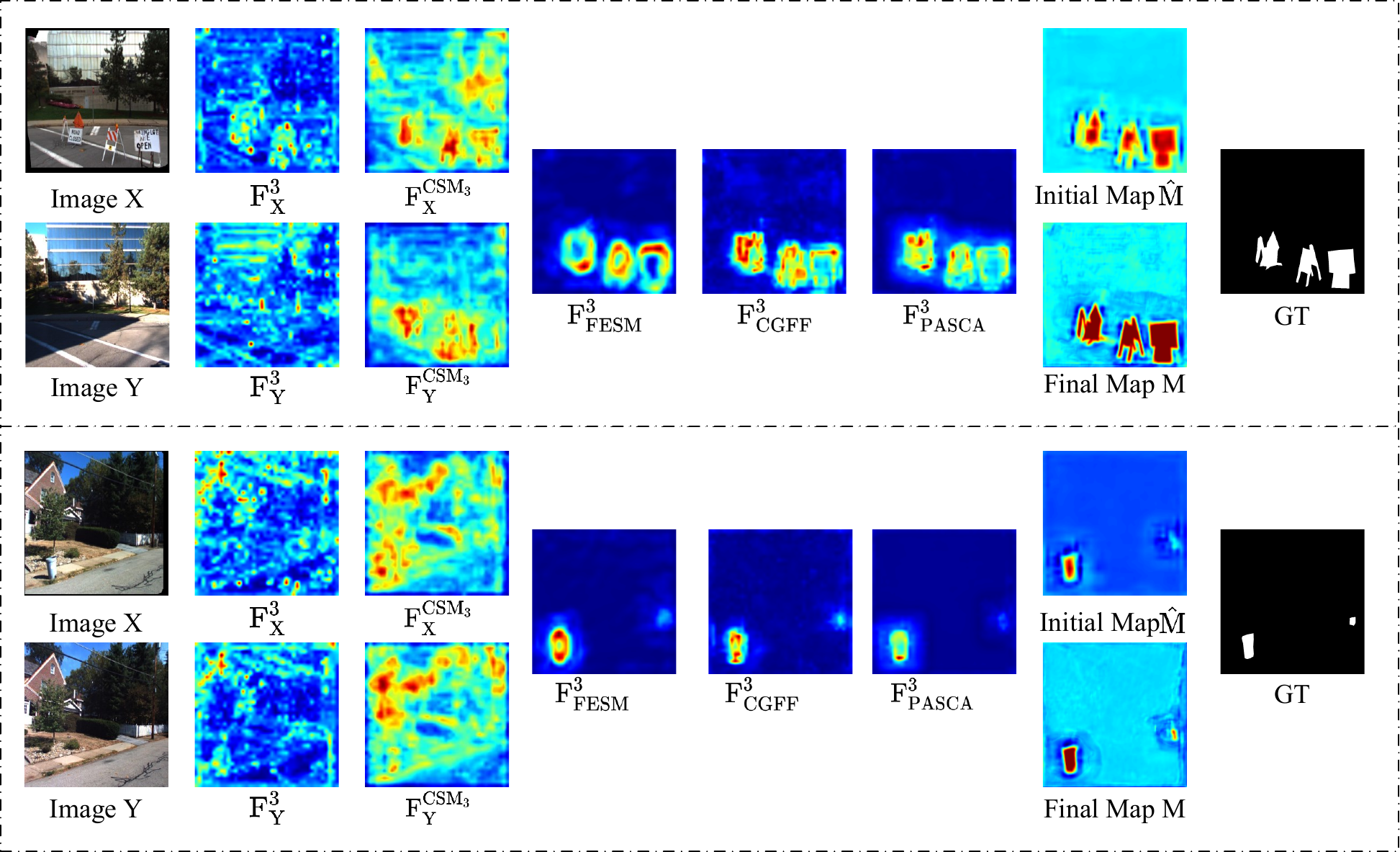} % 建议宽度占页面90%（预留页边距）
    \caption{Progressive feature refinement through key modules: Visualization results showing the effects of CSM, FESM, CGFF, and PASCA on image features and final change map compared with the ground truth.}
    \label{fig:visual}
\end{figure*}
\subsubsection{Module-Wise Feature Visualization Analysis}  
Fig.~\ref{fig:visual} illustrates the progressive refinement of features through our key modules. The initial siamese features reveal significant inter-image discrepancies, which are effectively suppressed by the CSM while preserving structural coherence. For change feature refinement, FESM selectively amplifies subtle changes, while CGFF integrates complementary patterns, jointly enhancing boundary precision. Spatial Refinement is achieved through PASCA, which captures multiscale context and enables pixel-wise localization of changes at varying scales. The initial change map is refined via attention-guided edge sharpening and multiscale feature fusion, yielding final map \textbf{M} that closely align with ground truth \textbf{GT}, balancing holistic accuracy and granular fidelity.

%%%%%%%%%%%%%%%%%%%%%%%%%%%%%%%%%%%%%%%%%%
\section{Conclusion}
\label{conclusion}
In this paper, we propose ‌CEBSNet‌, a novel change detection framework designed to mitigate the critical challenges of noise interference and subtle-change neglect in complex scenarios. By integrating the ‌Channel Swap Module‌, our method effectively models temporal dependencies between image pairs, reducing inter-pair discrepancies and enhancing robustness against illumination, seasonal variations, and other noise factors. The ‌Feature Excitation and Suppression Module‌ further improves detection accuracy by emphasizing change-related regions while suppressing background distractions. Additionally, the ‌Pyramid-Aware Spatial-Channel Attention module leverages multi-scale receptive fields and spatio-channel attention to capture changes of diverse sizes, ensuring comprehensive detection performance.
Extensive experiments on three common street-view and two remote sensing datasets demonstrate that CEBSNet outperforms sixteen state-of-the-art methods. Future efforts will focus on lightweight deployment strategies to balance computational efficiency and detection accuracy.

\subsection*{Authorship contribution statement}
Conceptualization, Q.X. and Y.X.; methodology, Q.X. and Y.X.; software, Q.X. and J.H; validation, Q.X. and J.H.; formal analysis, Q.X.; investigation, Q.X. and R.H.; resources, Q.X. and Y.X.; data curation, Q.X.; writing---original draft preparation, Q.X. and Y.J.; writing---review and editing, Q.X., Y.X. and R.H.; visualization, Y.J.; supervision, Y.X.; project administration, R.H.; funding acquisition, Y.X. and R.H.. All authors have read and agreed to the published version of the manuscript.

\subsection*{Funding}
This research was funded by the Scientific Research Program of Tianjin Municipal Education Commission 2023KJ232 and the Tianjin Natural Science Foundation General Project 24JCYBJC00990.

\bibliographystyle{plainnat}
\bibliography{reference}

\end{document}